\documentclass[10pt,twocolumn,letterpaper]{article}

\usepackage{wacv}
\usepackage{times}
\usepackage{epsfig}
\usepackage{graphicx}
\usepackage{amsmath}
\usepackage{amssymb}
\usepackage{booktabs}
\usepackage[accsupp]{axessibility}
\newcommand\kevin[1]{\textcolor{black}{#1}}
\newcommand\ze[1]{\textcolor{black}{#1}}

\def\wacvPaperID{1258} % Enter the WACV Paper ID here

\wacvalgorithmstrack   % Uncomment this line if you are submitting to the Algorithms Track.
\wacvfinalcopy % *** Uncomment this line for the final submission

\ifwacvfinal
\usepackage[breaklinks=true,bookmarks=false]{hyperref}
\else
\usepackage[pagebackref=true,breaklinks=true,colorlinks,bookmarks=false]{hyperref}
\fi

\begin{document}
\title{EventPoint: Self-Supervised Interest Point Detection and Description for Event-based Camera\\}

\author{Ze Huang\\
Xiamen University\\
\and
Li Sun\\
University of Sheffield\\
\and
Cheng Zhao\\
Bosch Research North America\\
\and
Song Li\\
Xiamen University\\
\and
Songzhi Su\thanks{Corresponding author: Songzhi Su, ssz@xmu.edu.cn}\\
Xiamen University \\
}

\maketitle
\thispagestyle{empty}

%%%%%%%%% ABSTRACT
\begin{abstract}

This paper proposes a self-supervised learned local detector and descriptor, called EventPoint, for event stream/camera tracking and registration.
Event-based cameras have grown in popularity because of their biological inspiration and low power consumption.
Despite this, applying local features directly to the event stream is difficult due to its peculiar data structure.
%Different from other existing local features methods for event streams, we propose a deep-learning-based method and training on high-resolution event format autonomous driving dataset, rather than simple or simulated situations.
We propose a new time-surface-like event stream representation method called Tencode. 
The event stream data processed by Tencode can obtain the pixel-level positioning of interest points while also simultaneously extracting descriptors through a neural network.
Instead of using costly and unreliable manual annotation, our network leverages the prior knowledge of local feature extraction on color images and conducts self-supervised learning via homographic and spatio-temporal adaptation.
To the best of our knowledge, our proposed method is the first research on event-based local features learning using a deep neural network.
We provide comprehensive experiments of feature point detection and matching, and three public datasets are used for evaluation (i.e. DSEC, N-Caltech101, and HVGA ATIS Corner Dataset).
The experimental findings demonstrate that our method outperforms SOTA in terms of feature point detection and description.
\end{abstract}

\section{Introduction}
\label{sec:intro}
In comparison with conventional standard frame-based cameras, the bio-inspired event camera offers significant advantages of microsecond temporal resolution, low latency, very high dynamic range, and low power consumption.
These revolutionary features enable some new robotics applications in extremely challenging conditions, e.g. in low-illumination scenarios and high-speed flying robot applications.
The event-based cameras, e.g. DVS~\cite{2008A}, Davis~\cite{brandli2014240} and ATIS~\cite{posch2010live} can capture the event points in the corresponding pixel position when sensing the pixel brightness changes over a temporal resolution.
More precisely, the event camera asynchronously measures changes in brightness of each pixel within a certain threshold in a high dynamic range from 60 $dB$ to 140 $dB$.
\kevin{The sign of events (positive or negative) is also known as polarity.}  

Although event-based cameras have numerous advantages, dealing with some standard computer vision tasks directly on the event stream, e.g. local feature extraction, is challenging due to the spatio-temporal data structure.
Local feature detection and representation~\cite{penev1996local}, is the core technology for a variety of applications, e.g. visual odometry, place recognition, 3D reconstruction, etc.
The image-based local feature extraction and description can be grouped into hand-crafted~\cite{lowe2004distinctive, calonder2010brief, rosten2006machine} and deep-learned~\cite{choy2016universal, yi2016lift, detone2018superpoint, simo2015discriminative, revaud2019r2d2} methods.
Compared to hand-crafted features, the deep-learned features demonstrate significant advances in terms of performance on several benchmarks~\cite{schonberger2017comparative, jin2021image}. 
%As the mainstream method in current academic circles, deep learned methods have been shown to be superior to hand-crafted methods~\cite{schonberger2017comparative}.

Local feature extraction methods on image data cannot be applied to event-based data straightforwardly due to the domain variance between the traditional image and event-based data.
Furthermore, the great sensitivity of the event camera creates a lot of noise, making the work more challenging~\cite{baldwin2020event, wang2020joint}.
Some recent research~\cite{vasco2016fast, mueggler2017fast, alzugaray2018asynchronous, manderscheid2019speed, benosman2013event} explore the corner points detection on the event stream. 
However, most of them only include interest point detection without describing the detected feature points due to the monotonic event-based data structure.
Some further event-based local descriptor methods~\cite{ramesh2019dart, clady2017motion} are evaluated using the toy or simulated event data, while the effectiveness, and robustness in large-scale realistic scenes have not been verified.
Additionally, the existing methods are handcrafted rather than deep-learned, which shows weakness in noise filtering, semantic-level understanding and adaptation of sensing data and hyper-parameters. 
%Certain preprocessing procedures, i.e. noise filtering are required since these event-based local descriptors are typically not robust to noise.
%In addition, almost all methods proposed are based on hand-craft, which often depend on hyperparameters change to adapt the input and ignore the powerful advantages of deep learning methods.

\begin{figure*}[htbp]
\centerline{\includegraphics[width=1.0\textwidth]{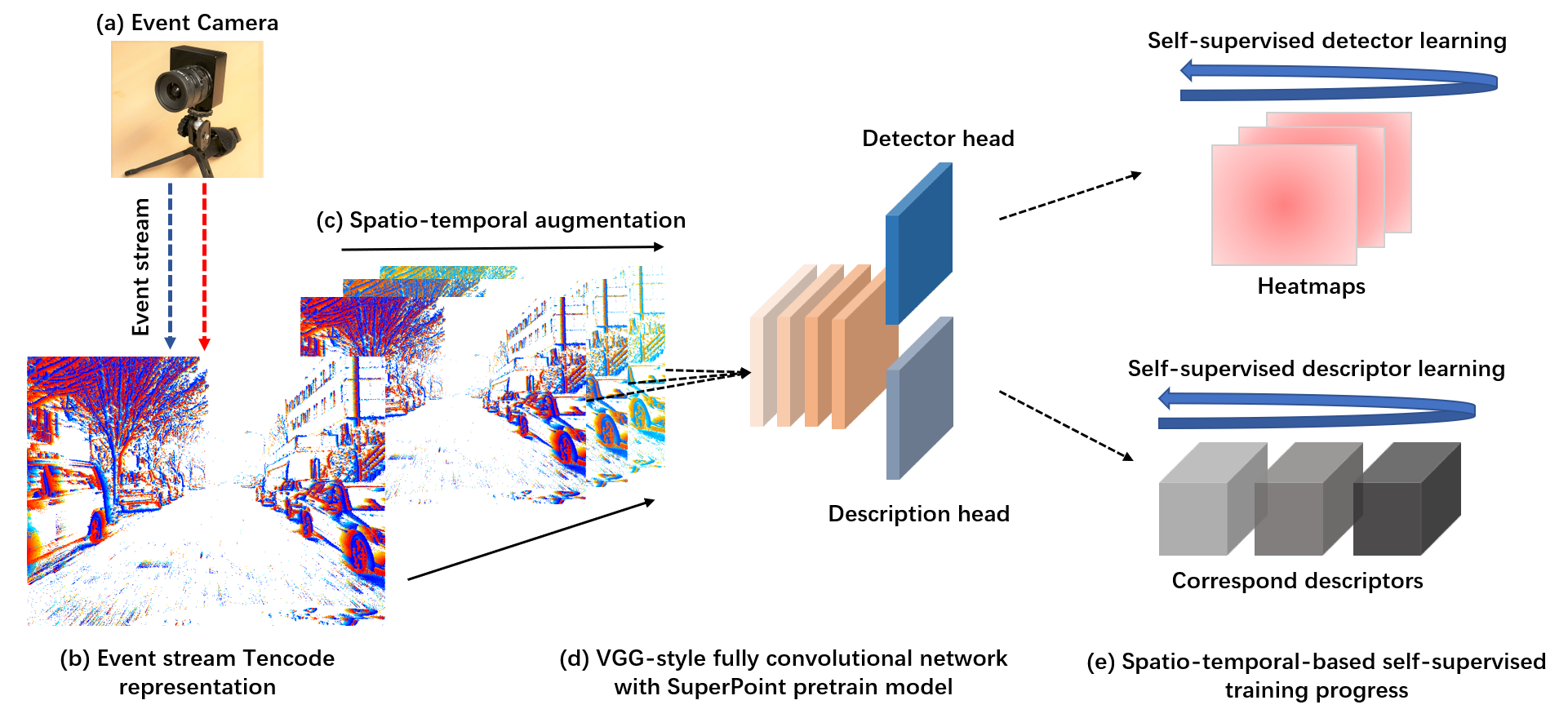}}
\caption{\textbf{Overview of our proposed method:} (a) event camera captures asynchronous event stream with binary polarity; (b) the event stream is sliced temporally and represented via Tencode; (c) spatio-temporal adaptation is used to generate the supervisory signal required for neural network training; (d) EventPoint uses SuperPoint-like architecture for local feature extraction; (e) two decoder heads, i.e. detector head and descriptor head, are trained separately in our proposed self-supervised manner.}
\label{figpipeline}
\end{figure*}

To overcome the limitations above, our research makes the following contributions:
\begin{itemize}
\item We propose a deep-learning-based local detector and descriptor, i.e., EventPoint, tailored for event stream/camera. 
\item We propose a simple but effective events representation method, called Tencode, which significantly facilitates the feature point's representation learning.
\item Our approach is delicately designed to learn spatio-temporal and homography invariant local descriptors in a self-supervised way without extra human annotations.
\item We conduct a comprehensive evaluation in terms of feature detecting and feature matching on three different public benchmarks, and the experimental results show that our approach is superior to existing methods.
\end{itemize}

%The remainder of this paper is organized as follows.
%Section~\ref{sec:rw} reviews related work on local feature on both RGB and event-based data. 
%Frequently used event stream representations are introduced in Section~\ref{sec:esr}.
%Tencode representation and the spatio-temporal augmentation we use are also introduced.
%Network architecture and the self-supervised learning we proposed are presented in detail in the rest of Section~\ref{sec:methods}.
%Section~\ref{sec:exp} shows the qualitative and quantitative results of our proposed method and other state-of-the-art methods on three evaluation experiments.
%Finally, in Section~\ref{sec:conc}, we give our conclusions.

\section{Related Work}
\label{sec:rw}
%related work这部分每一条引用基本都有修改
\subsection{Local Feature on Image Data}
\label{sec:lfrd}
%The conventional hand-crafted local features detect and describe the interesting points using the image geometric information.
%The most widely-used local feature, SIFT~\cite{lowe2004distinctive} achieves the scale and rotation invariance of local patch description. %which usually requires high computation on CPU.
%Based on SIFT~\cite{lowe2004distinctive}, the further proposed SURF~\cite{bay2008speeded} improves the run-time performance.
%A fast local feature, ORB~\cite{rublee2011orb}, is proposed which uses the extremely fast corner detection FAST~\cite{rosten2006machine} and a modified binary descriptor BRIEF~\cite{calonder2010brief}.

The hand-craft-based local features~\cite{lowe2004distinctive, bay2008speeded, rublee2011orb} are well-studied and are still the priority in real industrial applications.

Recently, the emerging deep-learned-based local features have become the dominating steam of methods.
LIFT~\cite{yi2016lift} is an early-stage work that firstly investigates the use of CNN to extract local features with a full pipeline of detection, orientation estimation, and feature description in a unified manner.
A lightweight deep-learned local descriptor, SuperPoint~\cite{detone2018superpoint} proposes a self-supervised learning method using pseudo-ground-truth correspondences generated by homographic transformation.
It designs a dual-head network for interested point detection and description separately.
Unsuperpoint~\cite{christiansen2019unsuperpoint} proposes a siamese network to learn the detector and descriptor, the interest points' positions are learned in a regression manner. 
%enabling interest point scores and positions to be learned automatically with a self-supervised approach.
\ze{R2D2~\cite{revaud2019r2d2} learns both keypoint repeatability and a confidence for interest points reliability from relevant training data, where style transfer method is used to increase robustness against dynamic illumination change such as day-night.}
By leveraging implicit semantic understanding, the learning-based local features~\cite{yi2016lift, detone2018superpoint, christiansen2019unsuperpoint, revaud2019r2d2} show extraordinary advances in dealing with long-term variation in \ze{real-world conditions~\cite{schonberger2017comparative, jin2021image}.}

%\subsection{Self-Supervised Learning for Event-based Data}
%\label{sec:sled}
%Self-supervised learning leverages temporal or geometry consistency as auxiliary tasks to provide supervision, therefore large-scale manually annotated data is not required.
%Compared with RGB-based vision dataset, event-based datasets are extraordinarily rare, and few self-supervised learning approaches on event-based data is available.
%Some recent research explore the optical flow estimation on event-based data under self-supervised learning.
%Most of the existing works using self-supervised learning on event data focus on optical flow.
%Self-supervised learning is suitable for event-based tasks which lacks annotated datasets, and most of the existing works using self-supervised learning on event-based data are focus on optical flow.

%EV-FlowNet~\cite{zhu2018ev} proposes a self-supervised photometric consistency loss to learning optical flow on frame-based events.
%It was further expanded to challenge depth estimation and egomotion in an unsupervised manner~\cite{zhu2019unsupervised}.
%In~\cite{paredes2021back}, a self-supervised approach incorporates both contrast maximization proxy loss and the event-based photometric consistency to learn optical flow estimation and image reconstruction from event-based data.
%The method~\cite{hagenaars2021self} focus on the complex task of learning to estimate optical flow from event-based camera inputs in a self-supervised manner.
%It modifies the spiking neural network training pipeline to encode minimal temporal information in the inputs.

\subsection{Local Feature on Event-based Data}
\label{sec:lfed}
%写出五种最好的event local feature文献,每个文献1-2句话
%Here, we focus works on local feature extraction based on pure event-based data, and those works based on simulated event data or multi-modality data, e.g. \cite{chiberre2021detecting} are not included.
Recently, the event-based local feature has been attracting a lot of attention from the computer vision community. 
EvFast~\cite{mueggler2017fast} employs the FAST corner point detection~\cite{trajkovic1998fast} to select the interesting event points via timestamp difference.   
%which regards those event points whose timestamp are significantly higher than other events within a certain radius as feature points.
EvHarris~\cite{vasco2016fast} transforms the raw event stream to a Time-surface~\cite{benosman2013event} representation, and further detects interesting points by Harris corner detector~\cite{derpanis2004harris}.
A more efficient Harris corner detector~\cite{glover2021luvharris} on event-stream is designed by tuning throughout of Time-surface and refactoring the conventional Harris algorithm.
In~\cite{manderscheid2019speed}, a random forest is employed to extract corner interesting point, and, Speed-invariant Time-surface feature is used for training. 
%In the research~\cite{manderscheid2019speed}, a random forest based on learning combining with the Speed-invariant Time-surface are proposed for event-based feature point detection, event points will be judged whether it is a corner point by binary classification.
The above methods only achieve event-based local feature detection, however, local feature description is not considered.
DART~\cite{ramesh2019dart} uses the log-polar grid of the event camera to encode the structure context to describe the interesting points.
Currently, most of the event-based local features require elaborate human design, which shows limited ability to handle complex situations such as significant motion changes or high-speed motion.   
%In summary, most of the methods for feature point extraction on event streams are manual-based and draw on traditional methods on images.
%Those methods~\cite{mueggler2017fast}~\cite{vasco2016fast}~\cite{alzugaray2018asynchronous} show weakness when dealing with complex situations such as motion change, or high-speed motion.
Most recent research~\cite{manderscheid2019speed} learns a local feature from the event data stream, but a large-scale human annotation is required.   

\section{Methodology}
\label{sec:methods}
%In this section, we first introduce the proposed event-stream representation, named Tencode, then we introduce the architecture of EventPoint following the detection and description learning.
An overview of our method is given in Fig.~\ref{figpipeline}.  
\subsection{Event Stream Representation}
\label{sec:esr}
\kevin{Event-stream data consists of four dimensional information $(x,y,t,p)$, where $x$, $y$ are the event's location, $t$ for the timestamp, and $p$ for the event's polarity.
The mainstream event stream representation methods can be grouped into two categories, i.e., Time-window-based representation and Time-surface-based representation.}

\kevin{Time-window-based methods use a constant temporal resolution $\Delta t$, then accumulate all events under a time window $(T, T+\Delta t)$ to generate a single frame representation $F$.}
\begin{equation}
F[x,y]=p\leftarrow (x,y,t,p),
\end{equation}
%where $(x,y,t,p)$ means a single event and $[x,y,p]$ means the information maintained by the Time-window-based representation. 
Time-window-based representation mainly considers the polarity of events but ignores the timestamp information. i.e., the asynchrony of event-based data, which is frequently used in global feature extraction or multi-modality fusion.
\kevin{The Time-surface-based representation~\cite{benosman2013event} transforms the spatio-temporal event stream into a frame-like representation $F$ by normalizing the timestamps, however the events' polarity is ignored.}
\begin{equation}
F[x,y]=t\leftarrow (x,y,t,p).
\end{equation}
\kevin{Time-surface is widely-used in tasks that require accurate local information.}

\begin{figure}[htbp]
\centerline{\includegraphics[width=0.5\textwidth]{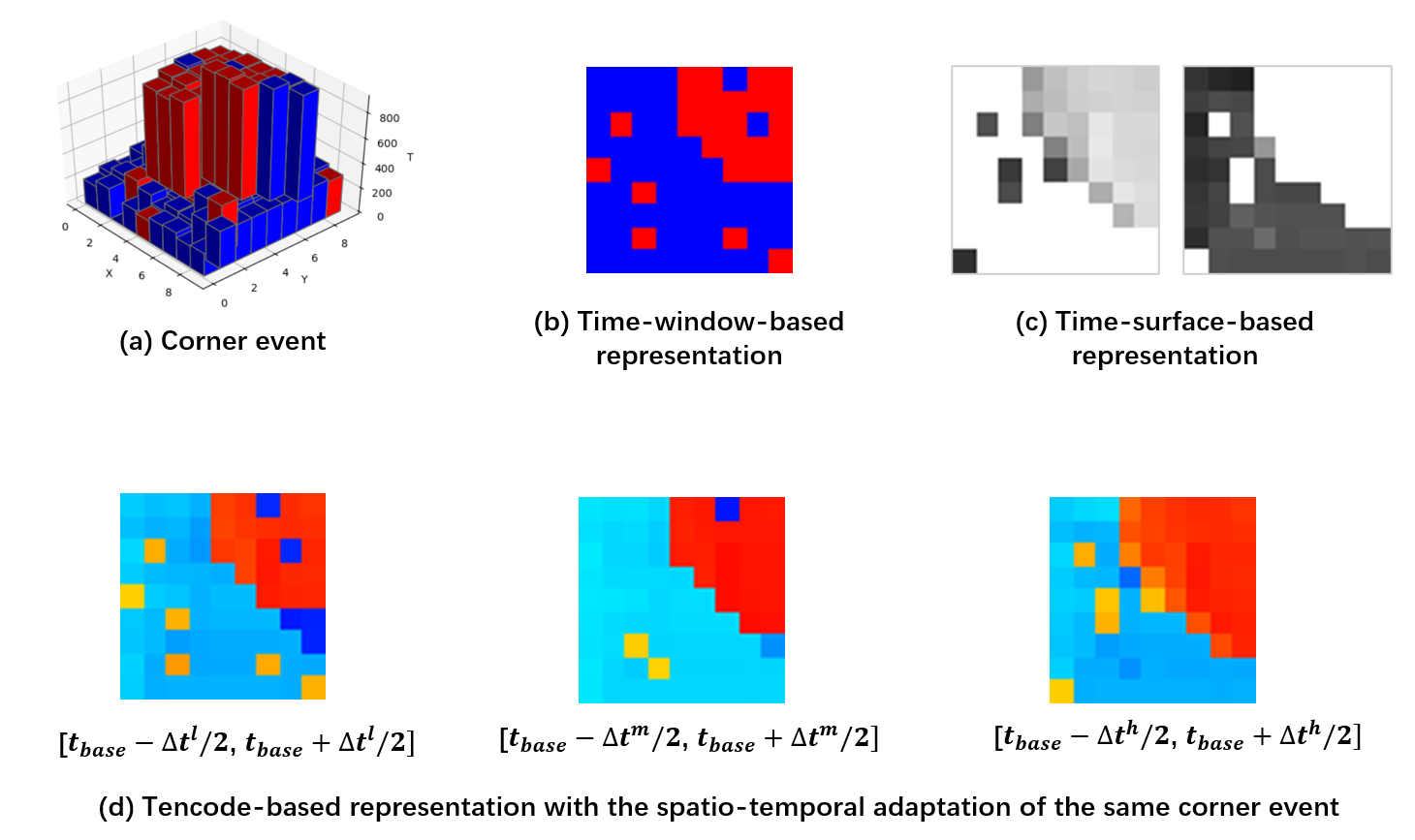}}
\caption{\kevin{\textbf{Visualization of different event-stream representation methods. A local patch of a corner is shown as an example:} (a) In a raw event stream, the corner point experiences a sudden change over timestamps; (b) Time-window-based representation maps events into a fixed temporal resolution to a single frame and only the polarity is presented (it ignores the magnitude of the time difference within this time window); (c) Time-surface-based representation captures the gradient of time information for each event, but either ignores the polarity or handles different polarities separately; (d) The proposed Tencode representation considers both the polarity and the gradient of timestamp.} 
%\ze{the purpose of spatio-temporal adaptation is to generate the consistency of the corner events under different temporal resolutions and the consistency is regarded as the self-supervisory signal.}
}
\vspace{0cm}
\label{figurepresentation}
\end{figure}

\kevin{Instead, we propose a new representation, named Tencode, taking both polarities and timestamps of event steam into account.
Firstly, a maximum temporal resolution $\Delta t$ is defined to discretize the continuous events to separate frames.}
%eliminate the impact of premature events on the representation of subsequent event streams.
Then, all events falling in this temporal window will be mapped into a frame $F$ according to polarity and timestamp information by,

\begin{equation}
\begin{aligned}
F[x,y] = (255,\frac{255*(t_{max}-t)}{\Delta t},0)&\leftarrow (x,y,t,+1),\\
F[x,y] = (0,\frac{255*(t_{max}-t)}{\Delta t},255)&\leftarrow (x,y,t,-1),
\end{aligned}
\end{equation}
\kevin{where $t_{max}$ represents the timestamp of the latest event in the temporal resolution $[t_{max}-\Delta t,t_{max}]$, and $+1$, $-1$ are positive and negative events respectively. }
\kevin{The three-dimensional vector $F[x,y]$ refers to the per-pixel information of the mapped frame $F$ at location $[x,y]$.
Following the expression of Time-window~\cite{chamorro2020high}, we set the first and third dimensions of the channel information to 255 or 0.} 
\ze{Specifically, 255 for the first channel and 0 for the third channel when dealing with events with positive polarity, similarly, 0 for the first channel and 255 for the third channel when dealing with events with negative polarity. }
\kevin{Fig.~\ref{figurepresentation} visualizes the two mainstream event stream representation methods (i.e. Time-surface and Time-window) and the proposed Tencode.}

\subsection{The Neural Network Input}
\label{sec:ni}
\kevin{Given a visual landmark that appears during a small temporal resolution $[T_{s}^{c}, T_{e}^{c}]$, we believe that any reasonable segment of the event stream $Ev$ within this temporal resolution can consistently detect the feature point and describe it,}
\begin{equation}
Ev[T_{s}^{1},T_{e}^{1}]\equiv Ev[T_{s}^{2},T_{e}^{2}],
\end{equation}
where $Ev[T_{s}^{n},T_{e}^{n}]$ refers to whole events with timestamps greater than $T_{s}^{n}$ and less than $T_{e}^{n}$.
% and $\equiv $ means equivalent to.
The constraint for this formula to hold is, $\delta_{max}>T_{e}^{i}-T_{s}^{i}>\delta_{min}$, $T_{e}^{i}<=T_{e}^{c}$, and $T_{s}^{i}>=T_{s}^{c}$.
\kevin{
We name this equivalence property the spatio-temporal consistency of the event stream.
$\delta_{max}$ and $\delta_{min}$ are the maximum duration and the minimum duration to hold the spatio-temporal consistency hypothesis.
We need this lower-bound to guarantee sufficient events accumulated to detect and describe the local feature.
While the upper-bound eliminates the random walk of the absence of the local descriptors.}
%The mapped representation is not enough to constitute feature points within a small temporal resolution.
%Conversely, the obvious appearance and disappearance of feature points occurs within a large temporal resolution.
%This is why we set upper and lower bounds.}
\kevin{The purpose of our method is to use a network to learn the spatio-temporal invariant feature point positions and descriptions given the above consistency hypothesis.
The physical meaning of this hypothesis is to achieve the speed-invariant representation of landmarks in the real world.}

%Due to the high dynamic range of the event camera, the spatio-temporal invariance can be approximated as the invariance of the corresponding moving feature points with different motion speeds photographed by a static event camera.

\kevin{Based on the above assumption, we used Tencode to generate $3$ event frames using different temporal resolutions as the neural network inputs. }%to the network and also as the supervision for self-supervised learning.
\kevin{For each base timestamp $t_{base}$, we choose a temporal resolution $\Delta t$ and get the latest event's timestamp $t_{max}$ in the temporal window $[t_{base}-\Delta t/2 , t_{base}+\Delta t/2]$.}
\kevin{The events in $3$ different temporal resolutions are further mapped via Tencode with parameter $t_{max}$ and $\Delta t$: }
\begin{equation}
\begin{aligned}
Ev[t_{base}-\frac{\Delta t^{h}}{2} , t_{base}+\frac{\Delta t^{h}}{2}]\xrightarrow[]{Tencode} &F_{h},\\
Ev[t_{base}-\frac{\Delta t^{m}}{2} , t_{base}+\frac{\Delta t^{m}}{2}]\xrightarrow[]{Tencode} &F_{m},\\
Ev[t_{base}-\frac{\Delta t^{l}}{2} , t_{base}+\frac{\Delta t^{l}}{2}]\xrightarrow[]{Tencode} &F_{l},
\end{aligned}
\end{equation}
where $\Delta t^{l}<\Delta t^{m}<\Delta t^{h}$.
\kevin{As shown in Fig.\ref{figurepresentation} (d), for the same patch, Tencode encodings of different temporal resolutions are distinct.
We leverage the neural network to learn invariant representations over these distinct Tencode encodings.}

\subsection{EventPoint Network Architecture}
\label{sec:arch}
\kevin{We employ the SuperPoint-like~\cite{detone2018superpoint} architecture consisting of a shared encoder and two heads, i.e., interest point detection and description.
The detailed architecture of the network is provided in Tab.\ref{tabnet}.}
\kevin{The VGG-style~\cite{simonyan2014very} encoder transforms the gray-scale Tencode representation $F\in \Re ^{H*W}$ to a low-resolution and high-dimensional feature map $f\in \Re ^{H/8*W/8*128}$.
Then the feature map is fed into two heads: one for interest points detection and the other one for description. 
The interest point detector head outputs a heatmap $h\in \Re ^{H/8*W/8*65}$ to give the probability of that pixel laying in an $8*8+1$ sized bin via a $Softmax$ function.
The last channel value represents whether the bin presences a feature points or not.}
\kevin{The heatmap $h$ will be further restored to the original size through the $Reshape$ operation,}
\begin{equation}
h\in \Re ^{H/8*W/8*65}\xrightarrow[65]{Softmax}\xrightarrow[]{Reshape}h_{out}\in \Re ^{H*W}.
\end{equation}
\kevin{The point's probability larger than a certain threshold $\tau$ is regarded as interest points.
The description head firstly outputs a dense grid of descriptors $d\in \Re ^{H/8*W/8*128}$, and then obtain a dense descriptor of the same size of the original frame through bi-cubic interpolation:}
\begin{equation}
d\in \Re ^{H/8*W/8*128}\xrightarrow[]{bi-cubic}d_{out}\in \Re ^{H*W*128}.
\end{equation}
\kevin{No deconvolution operation is used to guarantee the real-time performance.
In order to transferring the learned weights from pretained model, we use the same architecture with SuperPoint.}  
%Although we use the same architecture as SuperPoint for transfer learning with its pretrain model, 
\kevin{We further train the detector and descriptor via the spatio-temporal correspondences.}
%which is different from homography-induced correspondences.
%the spatio-temporally induced correspondence is the main driving force for our detector and descriptor learning.
%It is different from homography-induced correspondence in conventional training progress.

\begin{table}
\centering
%\captionsetup{labelformat=empty}
\caption{Detailed EventPoint Architecture}
\begin{tabular}{|l|c|l|c|} 
\hline
\multicolumn{4}{|c|}{\textbf{Encoder}}                                                                                                                                                                                                                        \\ 
\hline
\textbf{1a}   & \multicolumn{3}{c|}{ReLU(Conv2d(1,64))}                                                                                                                                                                                                       \\
\textbf{1b}   & \multicolumn{3}{c|}{ReLU(Conv2d(64,64))}                                                                                                                                                                                                      \\
              & \multicolumn{3}{c|}{MaxPool2d\textcolor[rgb]{0.055,0.067,0.086}{(}\textcolor[rgb]{0.055,0.067,0.086}{kernel\_size}=2\textcolor[rgb]{0.055,0.067,0.086}{, }\textcolor[rgb]{0.055,0.067,0.086}{stride}=2\textcolor[rgb]{0.055,0.067,0.086}{)}}  \\
\textbf{2a}   & \multicolumn{3}{c|}{ReLU(Conv2d(64,64))}                                                                                                                                                                                                      \\
\textbf{2b}   & \multicolumn{3}{c|}{ReLU(Conv2d(64,64))}                                                                                                                                                                                                      \\
              & \multicolumn{3}{c|}{MaxPool2d\textcolor[rgb]{0.055,0.067,0.086}{(}\textcolor[rgb]{0.055,0.067,0.086}{kernel\_size}=2\textcolor[rgb]{0.055,0.067,0.086}{,~}\textcolor[rgb]{0.055,0.067,0.086}{stride}=2\textcolor[rgb]{0.055,0.067,0.086}{)}}  \\
\textbf{3a}   & \multicolumn{3}{c|}{ReLU(Conv2d(64,128))}                                                                                                                                                                                                     \\
\textbf{3b}   & \multicolumn{3}{c|}{ReLU(Conv2d(128,128))}                                                                                                                                                                                                    \\
              & \multicolumn{3}{c|}{MaxPool2d\textcolor[rgb]{0.055,0.067,0.086}{(}\textcolor[rgb]{0.055,0.067,0.086}{kernel\_size}=2\textcolor[rgb]{0.055,0.067,0.086}{,~}\textcolor[rgb]{0.055,0.067,0.086}{stride}=2\textcolor[rgb]{0.055,0.067,0.086}{)}}  \\
\textbf{4a}   & \multicolumn{3}{c|}{ReLU(Conv2d(128,128))}                                                                                                                                                                                                    \\
\textbf{4b}   & \multicolumn{3}{c|}{ReLU(Conv2d(128,128))}                                                                                                                                                                                                    \\ 
\hline
\multicolumn{2}{|c|}{\textbf{Detector head}} & \multicolumn{2}{c|}{\textbf{Descriptor head}}                                                                                                                                                                \\ 
\hline
\textbf{cPa}  & Conv2d(128,256)               & \textbf{dDa}  & Conv2d(128,256)                                                                                                                                                                                 \\
              & ReLU                         &               & ReLU                                                                                                                                                                                           \\
\textbf{semi} & Conv2d(256,65)              & \textbf{desc} & Conv2d(256,256)                                                                                                                                                                                 \\
\hline
\end{tabular}
\label{tabnet}
\end{table}

\subsection{EventPoint Network Training}
\label{sec:enl}
\subsubsection{Detector Learning}
\label{sec:detec}
\begin{figure}[htbp]
\centerline{\includegraphics[width=0.5\textwidth]{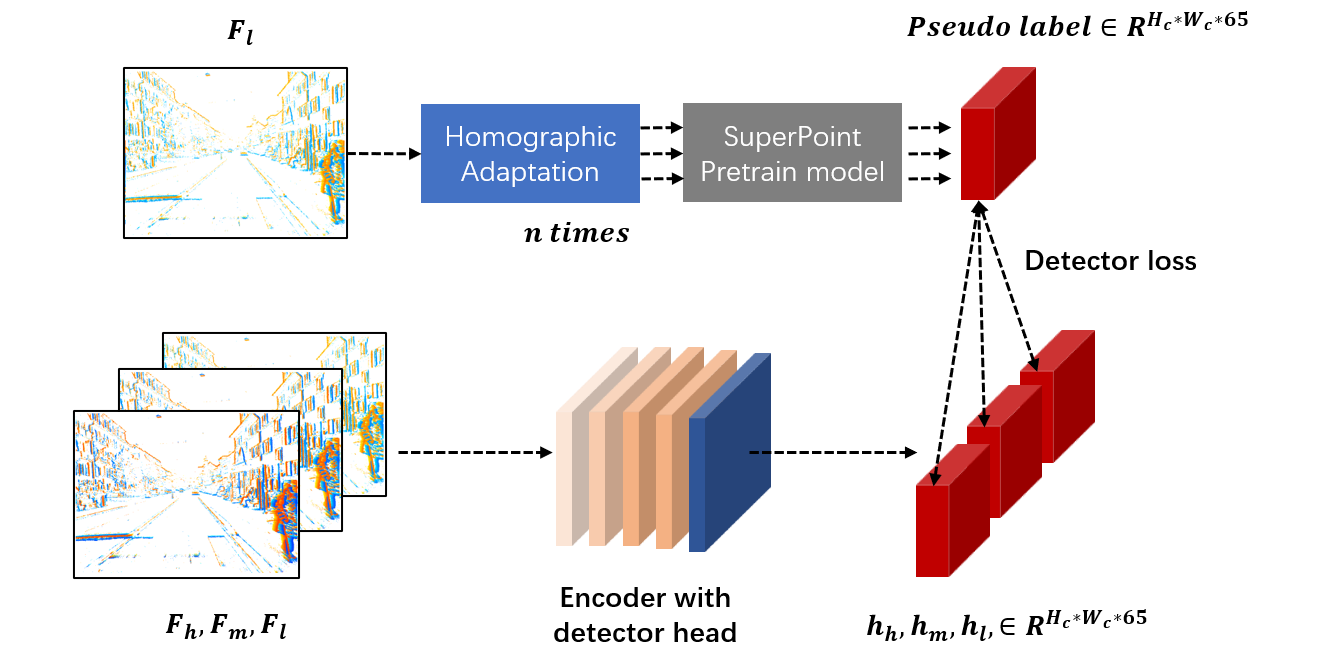}}
\caption{\kevin{\textbf{Spatio-temporal-based self-supervised detector learning}: we initially apply homographic adaptations to automatically annotate $F_{l}$ using SuperPoint's pretrained detector. Then, the inconsistency between the results of input and $label$ is penalized.}}
\label{figdetector}
\end{figure}

As mentioned in Sec~\ref{sec:ni}, we have $3$ Tencode frames $F_{h}$, $F_{m}$, and $F_{l}$ as inputs of the network.
In the detector learning step, the interest point position is firstly obtained by the detector of a pretrained SuperPoint on $F_{l}$ with the lowest temporal resolution.
Then more pseudo-label $label\in \Re ^{H/8*W/8*65}$ of interest point position is generated via homographic adaptation~\cite{detone2018superpoint}.
\kevin{To guide the detector training more effectively, we set predictions, of which probabilities are greater than a certain threshold $\tau$, as $1$ (positive labels) and other values to $0$ (negative labels).}
The detector learning pipeline is shown in Fig.~\ref{figdetector}.

\kevin{Since the number of detected interest points is much smaller than that of non-interest points, different from SuperPoint, we employ focal loss~\cite{lin2017focal} rather than cross-entropy loss to balance the training examples. }
The detection loss $Loss_{kp}$ is defined as:
\begin{equation}
Loss_{kp}= \sum_{t=h,m,l}w_{t}l_{p}(Softmax(h_{t}),label),   
\end{equation}
where $w_{t}$ refers to scale weights, and $h_{t}$ refers to the heatmaps generated through detector head of each Tencode frame input.
The detailed $l_{p}$ can be described as:
\begin{equation}
\begin{aligned}
l_{p}=\frac{1}{65H_{c}W_{c}}\sum_{i=1}^{H_{c}}\sum_{j=1}^{W_{c}}\sum_{k=1}^{65}Focal(h_{ijk},label_{ijk}),
\end{aligned}
\end{equation}
\kevin{where $H_{c}$ and $W_{c}$ are $1/8$ of the height and width of the original image resolution and the focal loss can be described as:}
\begin{equation}
Focal(\wp,\daleth)=\left\{\begin{matrix}\alpha (1-\wp)^{\gamma }ln(1-\wp) &\; \daleth=1
 \\ (1-\alpha) (\wp)^{\gamma }ln(\wp) &\; \daleth=0
\end{matrix}\right.
\end{equation}

\ze{where $\wp$ refers to the predict label, $\daleth $ refers to pseudo-label, $\alpha$ and $\gamma$ are hyper-parameters used to balance loss~\cite{lin2017focal}.}

\subsubsection{Descriptor Learning}
\label{sec:desc}

\begin{figure*}[htbp]
\centerline{\includegraphics[width=1.0\textwidth]{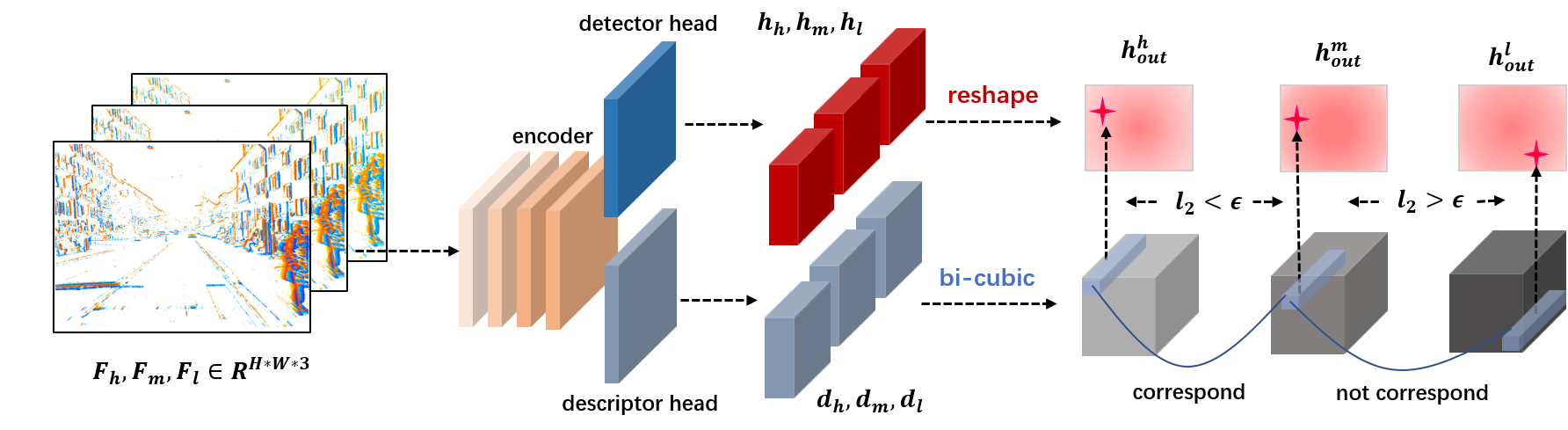}}
\caption{\kevin{\textbf{Spatio-temporal-based self-supervised descriptor training:} 
the association of detected interest point is decided according to their locations. 
Then the network learns to make the descriptor distance of correspondence pairs closer and that of non-correspondence pairs farther. }
%the basic criterion is that the corresponding descriptor pair get a closer distance, and other descriptor pairs get a longer distance. 
%The distance between the local feature points' positions provided by the trained detector head determines whether the descriptor pairs correspond or not.
}
\label{figtraindesc}
\end{figure*}

\kevin{After training the detector head, the descriptor head is further trained based on the detected interest points.
Firstly, the spatio-temporal correspondences of interest points between the Tencode frame triplets are defined as:}
\begin{equation}
\begin{aligned}
    s_{ij}^{i'j'}=\left\{\begin{matrix}
    1,&\quad L_{2}(ij,i'j') < \epsilon 
 \\ 0,&\quad otherwise
\end{matrix}\right.
\end{aligned}
\end{equation}
\kevin{Given an interest point position $(i_{h},j_{h})$ in $F_{h}$ and another one  $(i_{l},j_{l})$ in $F_{l}$.
If the Euclidean distance between $(i_{h},j_{h})$ and $(i_{l},j_{l})$ is less than a distance threshold $\epsilon$, we regard them as the correspond points, vice versa.}
The descriptor learning pipeline is shown in Fig.~\ref{figtraindesc}.

The $Hinge$-style loss is employed for the descriptor training:
\begin{equation}
Loss_{desc}= \sum_{t_{1},t_{2}=h,m,l}w_{t}l_{desc}(d_{t_{1}},d_{t_{2}}),   
\end{equation}
where $w_{t}$ refers to scale weights, and $d_{t}$ refers to the feature map of descriptors generated through descriptor head of each Tencode frame input.
The detailed $l_{desc}$ can be described as:
\begin{equation}
\begin{aligned}
l_{desc}=&\\
&\frac{1}{({H_{c}W_{c})}^{2}}\sum_{i,j=1}^{H_{c},W_{c}}\sum_{i',j'=1}^{H_{c},W_{c}}l_{d}(d_{t_{1}}^{ij},d_{t_{2}}^{i'j'};s_{ij}^{i'j'}),
\end{aligned}
\end{equation}
where $H_{c}$ and $W_{c}$ are $1/8$ of the height and width of the original image. And $l_{d}$ is defined as:
\begin{equation}
\begin{split}
\begin{aligned}
l{_{d}}(d,d^{'};s)=&\lambda*s*max(0,m_{p}-d^{T}d')\\&+(1-s)*max(0,d^{T}d'-m_{n}),
\end{aligned}
\end{split}
\end{equation}
where $m_{p}$ and $m_{n}$ refers to positive and negative margins respectively. $\lambda$ is used to balance the potential number of negative and positive correspondences.
%the fact that there are more negative correspondences than positive ones.

\section{Experiments}
\label{sec:exp}
We evaluate the EventPoint comparing with baselines in local feature detecting and matching tasks on three public datasets, i.e., DSEC~\cite{gehrig2021dsec, Gehrig3dv2021}, N-Caltech101~\cite{orchard2015converting} and HVGA ATIS Corner Dataset~\cite{manderscheid2019speed}.
\subsection{Network Training Details}
\label{sec:ntd}
EventPoint is trained under two event stream representation methods as shown in Fig.~\ref{figurepresentation}(b), and Fig.~\ref{figurepresentation}(d), and distinguished by $EventPoint$ and $EventPoint-T$ identification.
The basic model is trained on the DSEC dataset.
During training, $\Delta t^{h}$ and $\Delta t^{m}$ are randomly generated in a fixed temporal resolution, $50ms\geqslant \Delta t^{h} \geqslant 35ms$ , $35ms\geqslant \Delta t^{m}\geqslant 20ms$, and $\Delta t^{l}$ is set to $20ms$.
In the detection loss function, $\tau$ is set to $0.005$. 
The points with a value greater than $0.005$ in the heatmap are regarded as interest points in training and $0.015$ in testing.
The parameters $\alpha$ and $\gamma $ in focal loss are set to $0.75$ and $2.0$, respectively.
The weights $w_{h,m,l}$ in the loss function is set to $0.5$, $0.5$ and $1.0$ respectively.
The distance threshold $\epsilon$ is set to $8$.
The value of $\lambda$ used to balance in descriptor loss is $0.001$. 
The positive margin $m_{p}$ and negative margin $m_{n}$ are set to $1$ and $0.2$ respectively.

The network is implemented under PyTorch~\cite{paszke2019pytorch} framework. 
It is trained with batch sizes of 8 and each batch contains 3 event frames.
The SGD solver with default parameters of $lr= 0.001$ is used during training. 
The input size is cropped to $320 * 240$ for the network training. 
During inferring, the size is set to $640 * 480$ on the DSEC and HVGA ATIS Corner datasets.
But the size is set to $320 * 240$ when dealing with the N-Caltech101 dataset because of its low resolution.
The detector and descriptor branches are trained for around $10$ epochs respectively. 

For the run-time performance, the EventPoint takes about $0.1s$ to load the network and about $0.02s$ to process a single picture on an Intel(R) Core(TM) i9-9900KF CPU which achieves real-time performance similar to SuperPoint.

\begin{figure}[htbp]
\centerline{\includegraphics[width=0.5\textwidth]{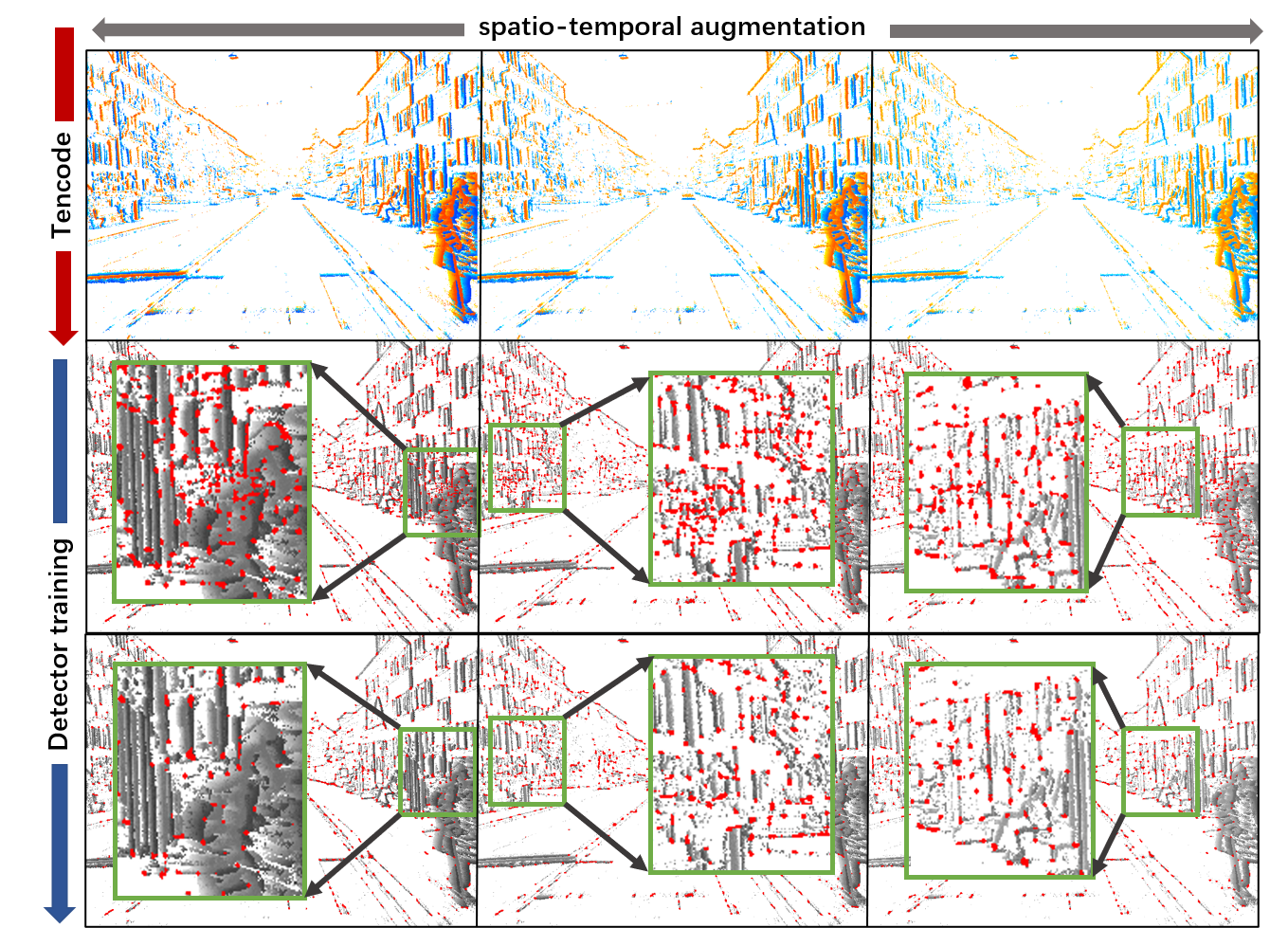}}
\caption{\textbf{Iterative detector training:} with the help of detector training, EventPoint can find more clear and more accurate interest points. Red dots represent the value greater than threshold $\tau$ in heatmaps. The red dots, i.e., interest points changes from large patches to stable positions gradually.}
\label{figvisdetector}
\end{figure}

\subsection{Datasets Details}
\label{sec:dd}
The DSEC~\cite{gehrig2021dsec, Gehrig3dv2021} dataset provides a set of sensory data in demanding illumination conditions.
Moreover, DSEC also provides the high resolution and large-scale stereo event camera dataset. 
It contains 53 sequences collected by driving in a variety of illumination conditions and provides ground-truth disparity map for the evaluation of event-based stereo algorithms.

The N-Caltech101 dataset~\cite{orchard2015converting} is the event format of the Caltech101 dataset~\cite{fei2004learning}. 
It consists of 101 object categories, each with a different number of samples ranging from 31 to 800. 
We followed the standard experimental setting of the dataset~\cite{lazebnik2006beyond} and~\cite{ramesh2017multiple}, which conducts feature matching evaluation by IOU matching score on up to 50 images in each category.

The HVGA ATIS Corner Dataset~\cite{manderscheid2019speed} is composed of 7 sequences of planar scenes acquired with an HVGA event sensor. 
We use the same evaluation metrics used in~\cite{manderscheid2019speed, chiberre2021detecting}, i.e., the reprojection error which is computed by estimating a homography from two different timestamps.

%@Henry，关于审稿人抱怨的SIFT没在RGB上比较的问题我们应该怎么修改，第一个实验中把sift去掉改成abalation study？ by huangze
\subsection{Ablation Study on DSEC}
\label{sec:elfd}
DSEC provides disparity maps corresponding to event frames under $50ms$ temporal resolution in urban-driving scenes. 
We selected several sequences with different brightness conditions to evaluate local feature matching quality via disparity map as a ablation study. 
The brightness conditions are directly related to the density of events and the number of noise. 
In the DSEC dataset, the number of events and noise in a dark environment is much more than that with better lighting conditions.

\begin{table}
\label{tab1}
\fontsize{8}{10}\selectfont
\caption{Ablation Study on DSEC dataset}
\centering
\begin{tabular}{|l|lll|} 
\hline
\multicolumn{4}{|c|}{zurich city 10\_b(dark)~ ~ ~ ~}\\ 
\hline
& $<3$ & $<6$ & $<9$ \\ 
\hline
EventPoint-T(ours) & 42.34 & \textbf{88.97} & \textbf{96.70} \\
EventPoint(ours) & \textbf{43.72} & 85.07 & 93.52 \\
Pretrained Weights\cite{detone2018superpoint}& 35.11 & 56.91 & 70.25 \\
\hline
\multicolumn{4}{|c|}{zurich city 11\_c(overcast)~ ~ ~ ~} \\ 
\hline
& $<3$ & $<6$ & $<9$ \\ 
\hline
EventPoint-T(ours) & 65.47 & \textbf{97.18} & \textbf{99.48} \\
EventPoint(ours) & \textbf{66.09} & 96.69 & 99.23 \\
Pretrained Weights\cite{detone2018superpoint}& 25.38 & 44.60 & 60.15 \\
\hline
\multicolumn{4}{|c|}{Inter laken 00\_c(sun)~ ~ ~ ~} \\ 
\hline
& $<3$ & $<6$ & $<9$ \\
\hline
EventPoint-T(ours) & \textbf{82.86} & \textbf{98.41} & \textbf{99.30} \\
EventPoint(ours) & 67.52 & 90.33 & 94.35 \\
Pretrained Weights\cite{detone2018superpoint}& 19.47 & 14.39 & 22.92 \\
\hline
\multicolumn{4}{|c|}{Inter laken 00\_d(sun)~ ~ ~ ~}\\ 
\hline
& $<3$ & $<6$ & $<9$ \\
\hline
EventPoint-T(ours) & \textbf{77.26} & \textbf{93.96} & \textbf{96.07}   \\
EventPoint(ours) & 67.46 & 91.88 & 94.61 \\
Pretrained Weights\cite{detone2018superpoint}& 19.47 & 35.04 & 46.91 \\
\hline
\end{tabular}\vspace{-1em}
\label{tabdis}
\end{table}

\begin{figure*}[htbp]
\centerline{\includegraphics[width=0.85\textwidth]{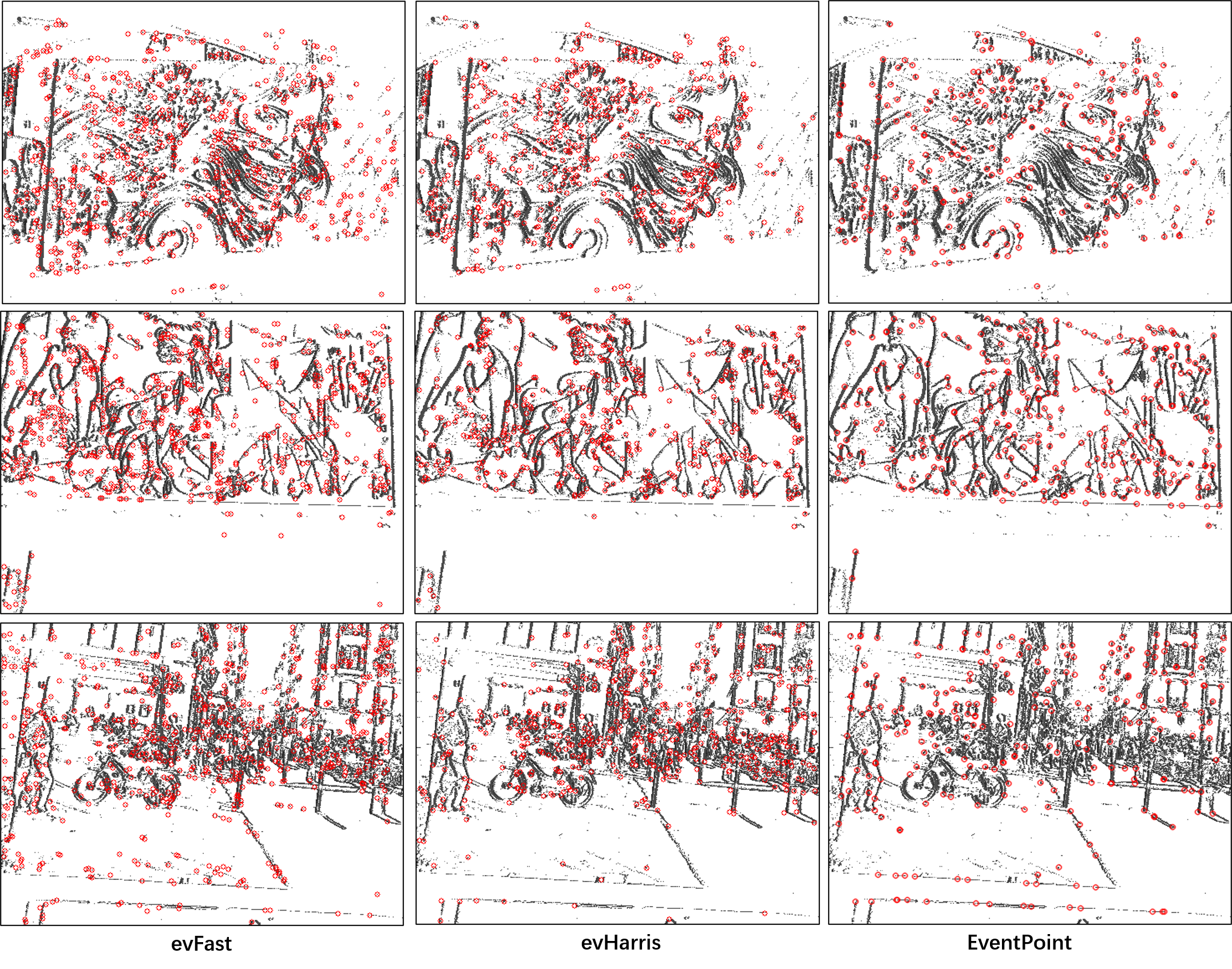}}
\caption{Qualitative results of feature point detection on HVGA ATIS Corner dataset with baselines. Events are mapped under 10ms temporal resolution with feature points detected. EventPoint can detect stable and accurate feature points and is more robust to noise.}
\label{figdetecvis}
\end{figure*}

The value $d$ of $(x, y)$ in the disparity map represents that the pixel points at the left frame $(x, y)$ correspond with the pixel points at the right frame $(x-d, y)$. 
For a matched pair $left(x, y)$, $right(x', y)$, if $x-x'< \sigma$, we regard it as a correct match.
The $\sigma$ is set to $3$, $6$ ,and $9$ in our experiment. 
The matching accuracy of each event data pair is calculated as the number of correct matches divided by the number of valid matches since the disparity map is relatively sparse.
The average matching precision on each sequence is reported as,
\begin{equation}
Precision = \frac{1}{N} \sum_{i=1}^{N} \sum_{j=1}^{M} \frac{\left [ \left | x - x' \right | < \sigma\right ] }{M},
\end{equation}
where $N$ refers to the number of samples in a single sequence, $M$ refers to correct matches in event data pairs, and $[\cdot]$ is the Iverson bracket.

From Tab.~\ref{tabdis}, it can be seen that the performance of the trained EventPoint significantly outperforms the initial network parameters, i.e., SuperPoint's pretrain weights.
In this ablation study, accurate position detection of feature points is an important step.
It shows our self-supervised detector training improves the detection performance.
We visualize the heatmaps' change during detector training in Fig.~\ref{figvisdetector}.
On the other hand, the proposed Tencode method introduces more rich timestamp cues into the training data to improve EventPoint's performance, especially in sunny times.

\subsection{IOU Matching Evaluation on N-Caltech101}
DART~\cite{ramesh2019dart} provides an evaluation metric, i.e., IOU matching score, for feature matching on the N-Caltech101 dataset. 
Given two event sequences with a length of about $300ms$, the global matching within the object contour is regarded as correct matching, otherwise as wrong matching.
To compare with work on descriptors on event streams, EventPoint is trained on N-Caltech101 and evaluated following the DART's experiment settings.

\begin{table}
    \centering
    \fontsize{8}{10}\selectfont
    \caption{Feature Matching IOU comparing with DART}
\begin{tabular}{l|c}
    \toprule
Methods & $IOU$\\
\hline
EventPoint-T(ours)  & \textbf{0.83} \\
EventPoint(ours)    & 0.79    \\
DART(FIFO size=2000)\cite{ramesh2019dart} & 0.72 \\
DART(FIFO size=5000)\cite{ramesh2019dart} & 0.67 \\
    \bottomrule
\end{tabular}\vspace{-1em}
    \label{tabiou}
\end{table}

\begin{figure}[htbp]
\centerline{\includegraphics[width=0.45\textwidth]{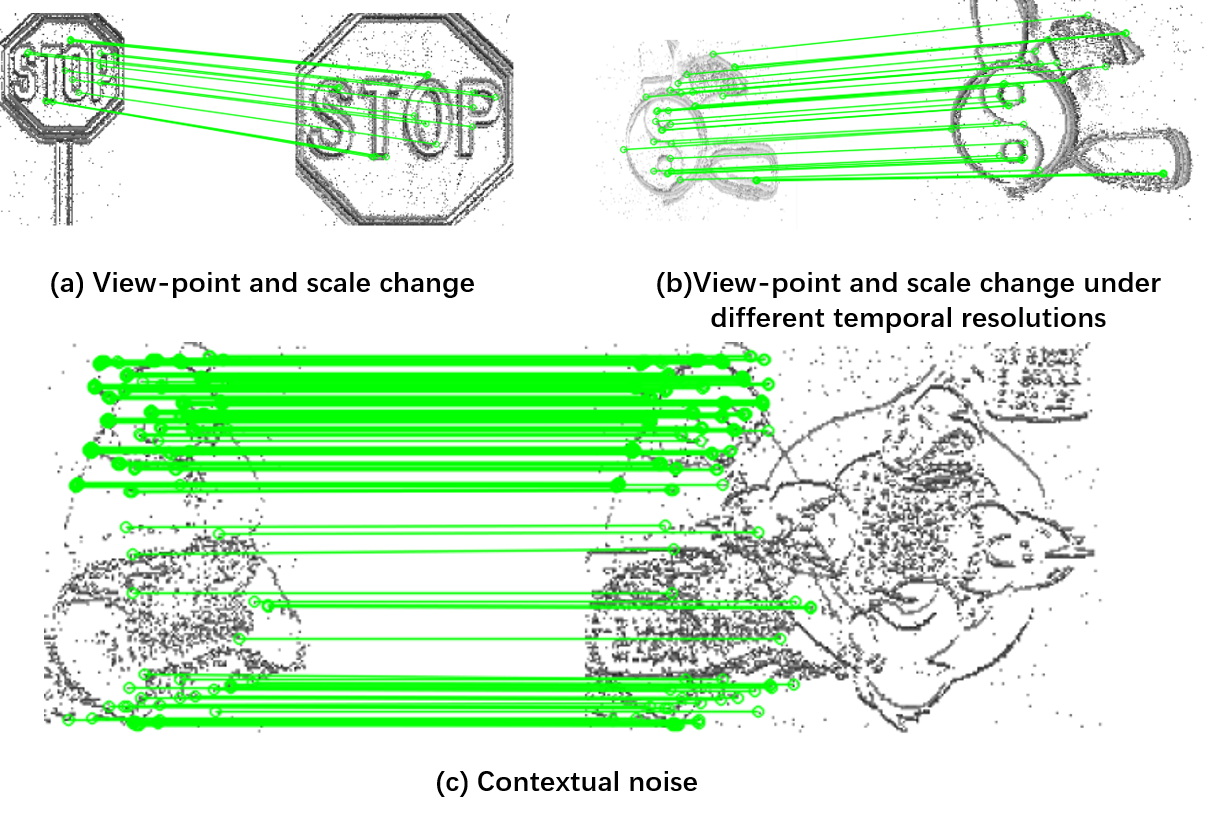}}
\caption{Qualitative results of feature matching on N-Caltech101. EventPoint is robust to view-point change, scale change, contextual background, and temporal resolution change.}
\label{figureNC}
\end{figure}

From Tab.~\ref{tabiou}, we can see the performance of EventPoint's deep-learning based description outperforms slightly the baselines DART's handcraft description.
It also shows the proposed Tencode method is more useful than the conventional encoding method even in single-objective-based datasets.
Fig.~\ref{figureNC} visualizes several samples of feature matching on N-Caltech101.

\subsection{Reprojection Evaluation on HVGA ATIS}
\label{sec:re}
HVGA ATIS Corner Dataset provides 7 sequences of planar scenes.
We compute reprojection error by estimating a homography between two different timestamps as used in \cite{manderscheid2019speed, chiberre2021detecting}.
In detail, given two timestamp $T_{1}, T_{2}$, and a temporal resolution $\delta t$,
we firstly use Tencode represents events fall $(T_{1}, T_{1}+\Delta t)$ and $(T_{2}, T_{2}+\Delta t)$ into two frames, then homographic transform is estimated by matching features of this two frames.
Once the homography is computed, we reproject points from time $T_{2}$ to the reference time $T_{1}$ and further compute the average distance between the reference points and the projected ones. 
The points detected outside the planar pattern are excluded.
We compare most of the current mainstream corner detection methods~\cite{vasco2016fast, mueggler2017fast, alzugaray2018asynchronous, manderscheid2019speed, chiberre2021detecting, benosman2013event}, as well as some image-based local feature extracting methods~\cite{lowe2004distinctive, detone2018superpoint} on the event stream.
Most of the existing works are limited to corner detection without description, or only the local region around the detected corner event is considered a description.
In this evaluation experiment, $\Delta t$ is set to $10ms$, and the margin of two timestamp, i.e., $T_{2} - T_{1}$ is set to $25ms$, $50ms$, $100ms$.
The DSEC and HVGA ATIS Corner datasets have the same resolution of $360 * 480$.
In order to verify the generalization ability of EventPoint, we train it only using the DSEC dataset but test on the HVGA ATIS Corner Dataset.
To prove the impact of different event stream representations, EventPoint is trained and tested under three different representations, i.e. Time-surface, Time-window, and the proposed Tencode. 
We use an OpenCV implementation, i.e, $findHomography()$ and $RANSAC$, with all the matches to compute the final homography estimation.

\begin{table}
    \centering
    \fontsize{8}{10}\selectfont
    \caption{Reprojection Error on HVGA ATIS Corner Dataset}
\begin{tabular}{l|l|ccc}
    \toprule
Methods & Representation & $25ms$ & $50ms$ & $100ms$ \\
\hline
%SIFT\cite{lowe2004distinctive} & Time-window & 12.56 & 35.09 & 48.20 \\
%SIFT\cite{lowe2004distinctive} & Tencode & 13.84 & 34.41 & 48.24 \\
evHarris\cite{vasco2016fast} & Time-surface & 2.57 & 3.46 & 4.58\\
Arc\cite{alzugaray2018asynchronous} & Time-surface &3.8 & 5.31 & 7.22 \\
evFast\cite{mueggler2017fast} & Time-surface & 2.12  & 2.63 & 3.18 \\
%SuperPoint\cite{detone2018superpoint} & Time-window &1.66 & 1.74 & 2.88 \\
%SuperPoint\cite{detone2018superpoint} & Tencode &1.47 & 2.46 & 2.85 \\
SILC\cite{manderscheid2019speed} &Speed-invariant & 2.45 & 3.02 & 3.68 \\
SILC\cite{manderscheid2019speed} &Time-surface & 5.79 & 8.48 & 12.26 \\
Chiberre et al.\cite{chiberre2021detecting} & Image gradients &2.56 & - & - \\
\hline
EventPoint(ours) &Time-surface & 1.46 & \underline{1.57} & \underline{1.89} \\
EventPoint(ours) &Time-window & \underline{1.41} & 1.61 & 2.39 \\
EventPoint(ours) &Tencode & \textbf{1.27} & \textbf{1.41} & \textbf{1.72} \\
    \bottomrule
\end{tabular}\vspace{-1em}
    \label{tabrepro}
\end{table}

From Tab.~\ref{tabrepro}, it can be seen that our method achieves the lowest reprojection error among all methods, and remains stable as the margin increasing between the two timestamps.
%We think the descriptor learning is the most helpful to improve the performance.
The experimental results show that our network learns a temporal representation invariance of corners on the event stream.
%including position and description.
Fig.~\ref{figdetecvis} visualizes the detection result comparing the baselines.
We use the proposed representation method Tencode with non-polarity separation avoiding the problem of detecting redundant corners at the polarity junction by the Time-surface-based methods. 
Thereby the reprojection error is further reduced.

\section{Conclusion}
\label{sec:conc}
In this paper, we present a novel self-supervised local feature EventPoint, including an interest point detector and a descriptor, for the event stream data.
We first represent the event stream under a temporal resolution by the proposed Tencode representation.
Then the EventPoint provides pixel-wise interest point locations and matches the corresponding descriptors from two dense Tencode representations. 
The proposed network is end-to-end trained in a self-supervised manner via homographic and spatio-temporal adaptation without expensive human annotation.
The experimental evaluations demonstrate that EventPoint achieves the SOTA performance of event feature point detection and description on DSEC, N-Caltech101, and HVGA ATIS Corner datasets.

\clearpage

{\small
\bibliographystyle{ieee_fullname}
\bibliography{egbib}
}
\end{document}